\documentclass{article}

\usepackage{microtype}
\usepackage{graphicx}
\usepackage{booktabs}
\usepackage{hyperref}

\makeatletter
\g@addto@macro{\UrlBreaks}{\UrlOrds}
\makeatother

\usepackage[accepted]{icml2026}

\usepackage{amsmath}
\usepackage{amssymb}

\usepackage{multirow}
\usepackage{enumitem}
\usepackage{tikz}
\usepackage{pgfplots}
\pgfplotsset{compat=1.18}
\usetikzlibrary{patterns}
\usepackage{float}
\usepackage{subcaption}
\usetikzlibrary{shapes.geometric, arrows.meta, positioning, patterns, calc, fit, decorations.pathreplacing}

\icmltitlerunning{Position: AI Leaderboards Are Underserving the Global South}

\newcommand{\position}[1]{\textbf{#1}}

\begin{document}

\twocolumn[
\icmltitle{Position: AI Leaderboards Are Underserving the Global South: A Case Study from India}

\begin{icmlauthorlist}
\icmlauthor{Sourav Banerjee}{iitkgp,shunya}
\icmlauthor{Saikat Saha}{nasscom}
\end{icmlauthorlist}

\icmlaffiliation{iitkgp}{Indian Institute of Technology Kharagpur, India}
\icmlaffiliation{shunya}{Shunya Labs, India}
\icmlaffiliation{nasscom}{Nasscom, India}

\icmlcorrespondingauthor{Sourav Banerjee}{sb.24@kgpian.iitkgp.ac.in}

\icmlkeywords{AI Evaluation, Leaderboards, Multilingual AI, Global South, India, Governance, Cultural AI, Accent Diversity}

\vskip 0.3in
]

\printAffiliationsAndNotice{}

\begin{abstract}
This position paper argues that AI leaderboards are structurally ill-suited to serving the Global South because they lack independent governance, conflict-of-interest policies, and mechanisms for metric evolution. The barrier is not missing data. High-quality regional benchmarks already exist: IndicSUPERB, MILU, and LAHAJA for India; IrokoBench for Africa; AlGhafa for Arabic. The barrier is institutional design. Global leaderboards do not include these benchmarks, and no governance mechanism compels them to do so. Commercial pressure corrects leaderboard failures when paying customers in the Global North are affected. The Global South lacks equivalent leverage. Without governance, failures affecting Hindi, Swahili, or Arabic speakers persist indefinitely as documented but unaddressed gaps. Using India as a case study (1.4 billion people, 22 scheduled languages, high-quality benchmarks but no trusted aggregation), we report findings from a consultation with 82 AI practitioners showing strong preferences for formal, non-government governance and for disclosure-based conflict management. Our contribution is institutional, not technical: we argue \textit{that} and \textit{why} governance is the load-bearing solution, with India as the worked case.
\end{abstract}

\section{Introduction}
\label{sec:introduction}

AI leaderboards are influential signals in the AI ecosystem. Governments reference them in procurement guidance \citep{omb2024ai,nature2025india}. Enterprises consult them for vendor selection. Investors cite them in due diligence. Research documents that companies invest heavily in benchmark performance, with estimates of hundreds of thousands of dollars in compute for top scores \citep{birhane2024trustbenchmarks}. When a leaderboard declares a model ``best,'' that signal enters procurement shortlists and investment memos. Because benchmarks define the implicit loss functions that guide model optimization, leaderboard governance shapes Machine Learning (ML) research and is a technical concern for the ML community.

\position{We argue that AI leaderboards, lacking conflict-of-interest (COI) policies and independent oversight, are structurally ill-suited to serving the Global South, and that regional leaderboard infrastructure with independent governance is necessary for emerging AI ecosystems to mature.}

\textbf{Terminology.} A \textit{benchmark} is a dataset paired with an evaluation protocol and metrics; MMLU is an example. A \textit{leaderboard} is a platform that aggregates benchmark results into rankings; the Open LLM Leaderboard is an example. By \textit{governance} we mean the institutional structures that determine how decisions are made, who makes them, and how conflicts are managed. A \textit{conflict of interest} (COI) exists when the same parties who benefit from high rankings also control the evaluation process, for example when a model developer also designs the benchmark, adjudicates edge cases, or decides when metrics are updated. This is not about malice. It is about structural incentives that can bias outcomes even with good intentions. Our critique targets leaderboard governance, not benchmark quality. Excellent benchmarks exist, but they lack the institutional infrastructure that would ensure their results are aggregated and ranked by parties without conflicts of interest.

Benchmark quality is not the bottleneck. High-quality regional benchmarks already exist: IndicSUPERB, MILU, and LAHAJA for India; IrokoBench for Africa \citep{adelani2025irokobench}; AlGhafa for Arabic. Global leaderboards simply do not use them. What is missing is \textit{institutional infrastructure}: trusted aggregation with transparent governance. The gap is institutional, not technical.

The stakes are immediate. India's \$1.2 billion IndiaAI Mission \citep{indiabudget2025} references global benchmarks that inadequately capture Indian linguistic reality. Similar patterns exist across the Global South: from Nigeria to Indonesia, governments and enterprises rely on leaderboard signals that do not reflect their populations' needs. A consultation with 82 AI practitioners (Section~\ref{subsec:consultation}) confirms demand for formal governance: every respondent endorsed some form of formal governance, 64\% preferred non-government stewardship, and 68\% favored disclosure-based conflict management over pre-emptive exclusion.

Our position is falsifiable: Section~\ref{sec:conclusion} specifies conditions under which regional infrastructure would be unnecessary. Throughout, we critique governance structures rather than specific organizations. The concentration of expertise across roles (model development, benchmark creation, leaderboard operation) is natural in nascent ecosystems. The question is whether governance structures exist to manage this concentration transparently.

\textbf{Scope and limitations.} As a position paper, this work argues \textit{that} regional leaderboard infrastructure with independent governance is necessary, not \textit{how} to implement it. We do not provide implementation blueprints, funding models, or technical specifications; these require context-specific deliberation beyond any single paper's scope. Our consultation (n=82) is illustrative, not representative; the sample skews toward production AI developers (52 of 82) and cannot speak for end users or civil society, the very groups we argue are underrepresented. Analogies we employ (the 10/90 gap, examination boards as discussed later) illuminate structural dynamics but have limits. We identify coordination challenges (federation across regional leaderboards) without solving them. These are directions for future work, not gaps that invalidate the core position.

\textbf{Conflict of Interest Disclosure.} S.\ Banerjee is a researcher at the Indian Institute of Technology Kharagpur and is also affiliated with Shunya Labs, an Indian voice-AI company. S.\ Saha is affiliated with Nasscom, the Indian industry association whose AI stakeholder networks distributed the practitioner consultation, part of which is reported in Section~\ref{subsec:consultation}. No Shunya Labs model is evaluated in this paper, and neither author serves on the governance board of any leaderboard discussed here. We disclose these affiliations in keeping with the governance norms this paper advocates.

\section{Why Leaderboards Matter}
\label{sec:why-leaderboards-matter}

AI leaderboards have evolved from academic scoreboards into critical infrastructure that shapes the AI ecosystem. Understanding who depends on them reveals the stakes of their limitations.

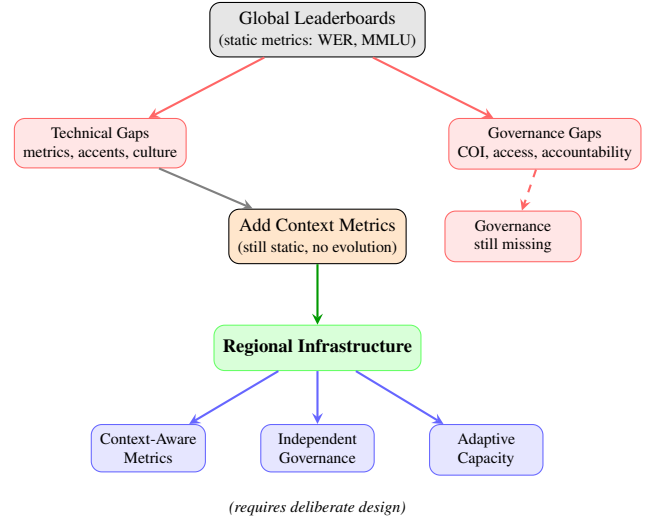
\begin{figure}[!htb]
\centering
\vspace{-0.5em}
\begin{tikzpicture}[
    node distance=0.5cm,
    state/.style={rectangle, draw, rounded corners, minimum width=2.2cm, minimum height=0.6cm, align=center, font=\scriptsize},
    problem/.style={rectangle, draw=red!60, fill=red!10, rounded corners, minimum width=1.8cm, minimum height=0.5cm, align=center, font=\tiny},
    solution/.style={rectangle, draw=green!60, fill=green!15, rounded corners, minimum width=2.4cm, minimum height=0.6cm, align=center, font=\scriptsize},
    requirement/.style={rectangle, draw=blue!60, fill=blue!10, rounded corners, minimum width=1.5cm, minimum height=0.5cm, align=center, font=\tiny},
    arrow/.style={->, thick, >=stealth}
]
\node[state, fill=gray!20] (current) {Global Leaderboards\\{\tiny (static metrics: WER, MMLU)}};

\node[problem, below left=0.8cm and 0.3cm of current] (tech) {Technical Gaps\\{\tiny metrics, accents, culture}};
\node[problem, below right=0.8cm and 0.3cm of current] (gov) {Governance Gaps\\{\tiny COI, access, accountability}};

\node[state, fill=orange!20, below=2.0cm of current] (partial) {Add Context Metrics\\{\tiny (still static, no evolution)}};
\node[problem, right=0.5cm of partial] (still) {Governance\\still missing};

\node[solution, below=0.8cm of partial] (proposed) {\textbf{Regional Infrastructure}};

\node[requirement, below left=0.7cm and 0.1cm of proposed] (p1) {Context-Aware\\Metrics};
\node[requirement, below=0.7cm of proposed] (p2) {Independent\\Governance};
\node[requirement, below right=0.7cm and 0.1cm of proposed] (p3) {Adaptive\\Capacity};

\node[font=\tiny\itshape, below=1.6cm of proposed] (label) {(requires deliberate design)};

\draw[arrow, red!60] (current) -- (tech);
\draw[arrow, red!60] (current) -- (gov);
\draw[arrow, gray] (tech) -- (partial);
\draw[arrow, dashed, red!60] (gov) -- (still);
\draw[arrow, green!60!black, thick] (partial) -- (proposed);
\draw[arrow, blue!60] (proposed) -- (p1);
\draw[arrow, blue!60] (proposed) -- (p2);
\draw[arrow, blue!60] (proposed) -- (p3);
\end{tikzpicture}
\vspace{-0.5em}
\caption{Evolution of the argument. Global leaderboards use static metrics with governance gaps. Adding context metrics helps but remains static and unaccountable. Regional infrastructure \textit{enables} but does not guarantee solutions: it must be deliberately designed with context-aware metrics, independent governance, and adaptive capacity.}
\label{fig:optimization-loop}
\vspace{-1em}
\end{figure}

\textbf{Leaderboards as implicit loss functions.} From an optimization perspective, leaderboards function as implicit loss functions for the AI development ecosystem. The metrics a leaderboard privileges become the objectives that organizations optimize against. When a leaderboard rewards Word Error Rate on English text, the optimization loop creates incentives that favor models excelling at English word boundaries but encoding representations poorly suited to agglutinative morphology. The deployment failures we document are consistent with systematic optimization pressures, not mere mismatches (Figure~\ref{fig:optimization-loop}).

\begin{table}[!htb]
\vspace{-0.5em}
\caption{Stakeholder Impact of Unreliable Leaderboards}
\label{tab:stakeholders}
\centering
\small
\begin{tabular}{@{}p{2cm}p{5cm}@{}}
\toprule
\textbf{Stakeholder} & \textbf{Impact in Multilingual AI Context} \\
\midrule
Governments & Select Automatic Speech Recognition (ASR) for citizen services using English-optimized metrics; deployed systems fail on regional dialects and code-switching\footnote{Alternating between two or more languages within a single utterance or conversation.} \\
\addlinespace
Enterprises & Choose customer service AI based on WER rankings that do not reflect local code-switching patterns (Hindi-English, Arabic-French, Spanglish) \\
\addlinespace
Startups & Cannot benchmark regional language specialization against global models; investor due diligence relies on irrelevant metrics \\
\addlinespace
Investors & May reference MMLU scores in due diligence; these do not predict regional language capability \\
\addlinespace
Researchers & Publish on benchmarks that reward English optimization; local language work appears ``lower performing'' \\
\addlinespace
End Users & Receive AI assistants that fail on agglutinative morphology, accents, and cultural context \\
\bottomrule
\end{tabular}
\end{table}

Table~\ref{tab:stakeholders} summarizes the stakeholder ecosystem. \textbf{Governments} reference global rankings in AI procurement and policy documents \citep[commentary]{nature2025india}. \textbf{Enterprises} cannot rely on English-centric leaderboards to predict code-switched performance, the dominant pattern in multilingual contact centers globally \citep{ai4bharat2023vistaar}. \textbf{Startups} building for local markets have no prominent venue to demonstrate regional language capabilities.

\textbf{Critically, ordinary citizens bear the risks but have no voice.} A farmer in rural India interacting with a government voice assistant, or a health worker in Nigeria using an AI diagnostic tool, consumes AI systems selected through leaderboard-influenced procurement, yet has no seat at governance tables. Research on ``algorithmic transference'' \citep{longoni2022transference} shows that faulty AI erodes trust in deploying institutions, not just technology. Technical and governance failures reinforce each other: inappropriate metrics persist because those who set them face no consequences from those who suffer the failures.

The consequences are documented. GPT-4 generates significantly more hallucinations in Hindi than English \citep{khanuja2025hallucination}; 88\% of AI-generated stories for Indian contexts contain cultural inaccuracies \citep{seth2025cultural}; leading ASR models perform significantly worse on minority dialects \citep{harris2024dialect}; agentic AI performance drops from 60\% under benchmark conditions to 25\% in production \citep{mehta2025enterprise}. These are systematic outcomes when models optimized for global benchmarks encounter regional reality.

\section{The Asymmetric Impact on Global South}
\label{sec:asymmetric}

Governance failures in AI leaderboards harm all regions but disproportionately impact the Global South. This section explains why.

\subsection{The Exclusion Problem}
\label{subsec:exclusion}

Global leaderboards do not merely have governance problems. They have content problems. The benchmarks they use systematically exclude the Global South.

Consider the HuggingFace Open ASR Leaderboard \citep{srivastav2025openasrleaderboard}\footnote{\url{https://huggingface.co/spaces/hf-audio/open_asr_leaderboard}, accessed 24 May 2026.} and its ``Multilingual ASR Evaluation'' section. It includes five languages: German, French, Italian, Spanish, and Portuguese. All five represent European language families, and the evaluation sets reflect European varieties rather than Brazilian Portuguese, West African French, or Latin American Spanish spoken by the majority of these languages' users. No Hindi (600M+ speakers), no Arabic (370M+ speakers), no Swahili (100M+ speakers), no Indonesian (200M+ speakers), no Bengali (270M+ speakers). Regional benchmarks exist for these languages: IndicSUPERB and LAHAJA for Indian languages, IrokoBench for African languages, AlGhafa for Arabic. The leaderboard simply does not use them.

This pattern extends beyond speech recognition. Research on MMLU found that 84.9\% of geography questions focus exclusively on North American or European regions \citep{globalmmlu2025}. English comprises 43.8\% of Common Crawl training data despite being spoken by less than 20\% of the world's population. Arabic, the fifth most spoken language globally, accounts for less than 1\% of training data. Over 2,000 African languages are largely neglected in AI models \citep{nature2025african}.

\textit{Data scarcity and governance failure are complementary, not opposed.} Upstream training-data imbalance (such as Arabic at less than 1\% of Common Crawl) is a real problem that requires investment in data collection and curation. The English share of the indexed web is itself roughly 50\%, so a 43.8\% share in Common Crawl reflects an internet-wide skew, not a benchmark design flaw. Our argument is about what happens \emph{after} such data exists. High-quality regional benchmarks (IndicSUPERB, MILU, LAHAJA, IrokoBench, AlGhafa) have been built but are still ignored by global leaderboards because no governance mechanism compels their inclusion. The data gap and the governance gap require different interventions, and both are necessary. Our paper addresses the second, not because the first is unimportant, but because it is already widely recognized while the governance failure is not.

\subsection{Why Markets Will Not Fix This}
\label{subsec:markets}

Global health R\&D governance produces systematic underinvestment in diseases affecting the Global South, not through malice but through structures optimizing for different priorities. The ``10/90 gap'' describes how only 10\% of health research addresses conditions causing 90\% of the global disease burden \citep{rottingen2013mapping}. Cancer research attracts funding because patients in wealthy countries can pay for treatments. Malaria research is underfunded because affected populations cannot.

AI evaluation infrastructure exhibits analogous neglect. When an English ASR model fails, enterprises escalate, contracts are at stake, and engineering resources are allocated to fix it. When the same model fails on Hindi or Hausa, the failure is documented in release notes as a scope limitation, acknowledged but not prioritized. The asymmetry is structural, not intentional.

\textit{This analogy to neglected tropical diseases is instructive only and has obvious limits}: AI models can be adapted through fine-tuning at lower cost than new drug trials \citep{xu2024finetuning}. But the analogy holds where it matters: fine-tuning \textit{capability} does not create fine-tuning \textit{incentives}. Without market pressure or governance mandates, technical possibility does not translate into actual adaptation.

Market pressure creates accountability for the Global North. It does not create accountability for the Global South. This is not a criticism of any organization. It is a description of structural incentives. No market force will push global leaderboards to adopt regional benchmarks for Santhali, Hausa, or Quechua speakers. Without governance mandating inclusion, inclusion will not happen.

\subsection{Four Mechanisms of Asymmetric Impact}
\label{subsec:mechanisms}

\textbf{Escape route asymmetry.} Sophisticated actors in resource-rich contexts can conduct independent evaluation or maintain internal benchmarking infrastructure. Resource-constrained actors must accept global signals as authoritative. A procurement officer in Bangalore must rely on leaderboards that a counterpart in Seattle can verify independently through enterprise-grade evaluation teams.

\textbf{Bug-fix priority asymmetry.} When GPT-4 hallucinates in English, enterprise customers file tickets and the issue receives engineering attention. When it hallucinates at higher rates in Hindi \citep{khanuja2025hallucination}, the behavior is characterized as expected performance variance for lower-resource languages. No institution currently has the mandate to treat Hindi performance degradation as a first-class defect.

\textbf{Institutional redundancy.} The Global North has multiple institutions checking AI capability claims: academic peer review, enterprise procurement teams, regulatory bodies, consumer advocacy organizations. The Global South often has single points of failure where a flawed leaderboard becomes the sole arbiter of quality.

\textbf{Optimization lock-in.} Leaderboards function as implicit loss functions, and architectural decisions favoring English compound over time as deployment generates training data. When leaderboards do not measure what matters to the Global South, models do not optimize for it. The gap widens.

\subsection{Governance as the Mechanism for Inclusion}
\label{subsec:governance-mechanism}

Where market forces are absent, governance structures become the primary pathway to inclusion. For the Global South, transparent leaderboard governance serves the function that commercial pressure serves for the Global North: it creates accountability. Without governance structures that mandate inclusion of regional benchmarks, regional languages, and regional contexts, inclusion will not happen. The data exists. Global leaderboards ignore it. Markets will not change this. Only governance can.

\section{Systematic Failures of Global Leaderboards}
\label{sec:failures}

Global AI leaderboards fail diverse regions through technical limitations compounded by governance failures that prevent correction. The technical problems are solvable. The governance problems are why they remain unsolved.

\subsection{Technical Failures}
\label{subsec:technical}

\textbf{Metric mismatch.} For ASR, Word Error Rate (WER) was designed for English and tends to disadvantage morphologically rich languages \citep{javed2023indicsuperb}. For LLMs, MMLU \citep{hendrycks2021mmlu} and similar benchmarks embed Western assumptions. Models ranking highly on MMLU show substantial degradation on India-specific questions (MILU, \citealt{ai4bharat2025milu}), African knowledge (IrokoBench, \citealt{adelani2025irokobench}), and Arabic contexts (AlGhafa, \citealt{alghafa2023}). The IndicParam benchmark \citep{indicparam2025} quantifies this gap: as reported in that benchmark, the best-performing model evaluated (Gemini-2.5) reaches only 58\% accuracy on low-resource Indic languages, with GPT-4 at 45\%, substantially below these models' English performance. \textit{A note on model vintage:} the reported numbers reflect models available at the time of the cited evaluations. The absence of frontier-model results on regional benchmarks is itself the governance gap we identify. No institution is mandated to keep these results current, and so even when newer models exist their behavior on Hindi, Yoruba, or Arabic remains unmeasured in any trusted public venue.

\textbf{Code-switching and accent blindness.} Code-mixing prevalence in multilingual societies reached 60\% by 2020 \citep{sengupta2024hinglish}, yet global benchmarks treat it as exceptional. The LAHAJA benchmark \citep{ai4bharat2024lahaja} reveals 15-30\% ASR performance degradation across Hindi regional accents. For LLMs, studies find significantly higher hallucination rates in Hindi and Farsi compared to English, even for questions rooted in Indic contexts.

\textbf{Knowledge conflicts.} Global models trained on Western corpora exhibit \textit{parametric dogmatism} \citep{conflictbank2024}: they prioritize training-time knowledge over retrieval-augmented local context, producing ``cultural hallucinations.'' A model may override Indian constitutional provisions with Western legal precedents, or contradict local medical practices even when explicitly provided as context. This is not a data quantity problem. It is a fundamental tension between model priors and local ground truth.

\textbf{The pattern extends across modalities.} We focus on language-model leaderboards because the governance gap is most documented there, but the same structural pattern holds for vision, medical, agricultural, and weather AI. Chest X-ray vision-language models underdiagnose marginalized groups, with the highest error rates for Black female patients \citep{yang2025chestxray}. Western cardiovascular risk models misclassified 80\% of 4{,}975 Indian first-time heart-attack patients as low/moderate risk \citep{gupta2026cardio}. PlantVillage crop-disease accuracy fell from 99.35\% on its North American benchmark \citep{mohanty2016plantvillage} to 49\% on Tanzanian cassava in deployment \citep{mrisho2020cassava}. AI weather models inherit ERA5 biases against tropical regions where Africa has one-eighth the WMO-recommended station density \citep{wmo2024stations,mozaffari2026weather}. In every case, no governance mechanism required validation on affected populations before deployment. Appendix~\ref{app:crossmodal} gives the full evidence.

Every technical problem above has a known fix. Better metrics exist. Regional benchmarks exist. The question is why global leaderboards have not adopted them. The answer is governance.

\subsection{Governance Failures}
\label{sec:governance}

\textbf{Structural COI in practice.} As defined in Section~\ref{sec:introduction}, a conflict of interest arises when the same parties who benefit from high rankings also control evaluation. Resource scarcity naturally concentrates expertise: the same people who have the skills to build models also have the skills to build benchmarks and run leaderboards. This concentration is expected in nascent ecosystems. The question is whether governance structures exist to manage it.

\begin{figure}[!htb]
\centering
\begin{tikzpicture}[
    node distance=0.6cm,
    topbox/.style={rectangle, draw=black!80, fill=green!20, rounded corners=2pt, minimum width=4.2cm, minimum height=0.7cm, align=center, font=\footnotesize, line width=0.6pt},
    actionbox/.style={rectangle, draw=black!80, fill=blue!15, rounded corners=2pt, minimum width=1.9cm, minimum height=0.65cm, align=center, font=\footnotesize, line width=0.6pt},
    centerbox/.style={rectangle, draw=black, line width=1pt, fill=white, rounded corners=2pt, minimum width=2.8cm, minimum height=0.7cm, align=center, font=\footnotesize\bfseries},
    issuebox/.style={rectangle, draw=black!70, fill=gray!18, rounded corners=2pt, minimum width=2.2cm, minimum height=0.55cm, align=center, font=\footnotesize}
]
\node[topbox] (contributor) {Major Compute Provider Employees};

\node[actionbox, below=0.8cm of contributor, xshift=-1.4cm] (submit) {Submit Models\\[-1pt]{\tiny (top-ranked)}};
\node[actionbox, below=0.8cm of contributor, xshift=1.4cm] (eval) {Evaluate Models\\[-1pt]{\tiny (own \& others)}};

\node[centerbox, below=1.9cm of contributor] (lb) {OpenASR Leaderboard};

\node[issuebox, below=0.65cm of lb, xshift=-1.4cm] (coi) {No COI Policy};
\node[issuebox, below=0.65cm of lb, xshift=1.4cm] (griev) {No Grievance Redressal};

\draw[-{Stealth[length=2mm]}, line width=0.8pt, black!70] (contributor) -- (submit);
\draw[-{Stealth[length=2mm]}, line width=0.8pt, black!70] (contributor) -- (eval);
\draw[-{Stealth[length=2mm]}, line width=0.8pt, black!70] (submit) -- (lb);
\draw[-{Stealth[length=2mm]}, line width=0.8pt, black!70] (eval) -- (lb);
\draw[-{Stealth[length=1.5mm]}, line width=0.6pt, black!50] (coi) -- (lb);
\draw[-{Stealth[length=1.5mm]}, line width=0.6pt, black!50] (griev) -- (lb);
\end{tikzpicture}
\caption{Structural pattern illustrated by the HuggingFace Open ASR Leaderboard: employees of a major compute provider who co-authored the leaderboard also develop top-ranked models, with no published COI policy or grievance redressal mechanism.}
\label{fig:openasr}
\vspace{-0.5em}
\end{figure}
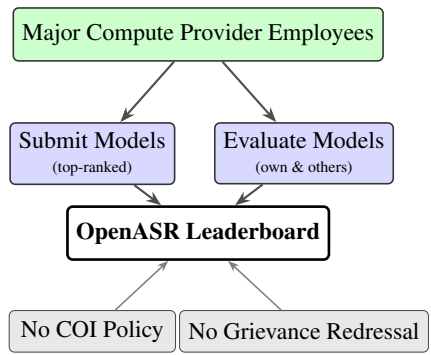

\textbf{The structural pattern.} Several prominent leaderboards allow the same players to submit models and evaluate competitors. The HuggingFace Open ASR Leaderboard illustrates this pattern (Figure~\ref{fig:openasr}): some co-authors of the leaderboard itself \citep{srivastav2025openasrleaderboard} also develop top-ranked models on that leaderboard \citep{sekoyan2025canary}, related training datasets by overlapping author teams \citep{koluguri2025granary}, and core architectural components \citep{rekesh2023fastconformer}, with no published COI policy or grievance redressal mechanism.

\textbf{Why transparency of methods is insufficient.} One might argue that if evaluation code is public, conflicts do not matter. But as in academic peer review, where criteria are public yet recusal is still required, the concern is not falsification but influence over \textit{which} metrics are chosen, \textit{how} edge cases are adjudicated, and \textit{when} benchmarks are updated. Transparency of \textit{execution} does not address conflicts in \textit{design}.

\textbf{Selective access.} \citet{singh2025leaderboard} documented that one provider evaluated 27 model variants privately before public release, and the top two providers received 39.6\% of arena evaluation data while 83 open-weight models combined received only 29.7\%. Unlike academic peer review, leaderboards have no formal disclosure requirements, no recusal processes, and no dispute resolution.

\textbf{Declining transparency.} The Foundation Model Transparency Index found that average transparency scores fell from 58 in 2024 to 40 in 2025 \citep{stanfordfmti2025}. Companies remain most opaque about training data, compute, and post-deployment usage. Transparency is getting worse, not better.

\textbf{Goodhart dynamics.} Goodhart's Law states that ``when a measure becomes a target, it ceases to be a good measure.'' AI leaderboards exhibit this dynamic at scale: models optimize for benchmark metrics, and the metrics progressively lose validity as measures of true capability \citep{thomas2022goodhart}. But Goodhart dynamics do not affect all populations equally. A metric designed for English word boundaries will eventually lose validity. But English speakers experience genuine capability improvements \textit{before} the metric breaks. Populations whose needs were never encoded in the metric experience only the costs of optimization (gaming, contamination, resource allocation to benchmark performance) without ever experiencing the benefits.

\section{What Institutional Infrastructure Means}
\label{sec:institutional}

The problem is not benchmarks. India has IndicSUPERB, MILU, LAHAJA. Africa has IrokoBench. The Arabic world has AlGhafa. Southeast Asia has SEA-HELM \citep{seahelm2024}. The data exists. The problem is that benchmarks without institutions are just datasets.

Consider what global leaderboards currently provide: a website, a ranking, and implicit trust that the operators are acting in good faith. What they lack is any structure that would \textit{justify} that trust. \textit{Institutional infrastructure} fills this gap:
\vspace{-0.5em}
\begin{itemize}[noitemsep,topsep=0pt,partopsep=0pt,parsep=0pt]
    \item \textbf{Trusted aggregation:} A neutral body that ranks models without conflicts of interest. Without this, rankings reflect the interests of whoever controls the leaderboard.
    \item \textbf{Governance:} Published COI policies, disclosure requirements, recusal processes. Without this, the same organization can create benchmarks, submit models, and declare winners.
    \item \textbf{Accountability:} Formal dispute resolution when rankings are contested. Without this, developers who believe they were unfairly evaluated have no recourse.
    \item \textbf{Evolution:} Mechanisms to adopt new metrics as understanding advances. Without this, benchmarks ossify even as the field learns their limitations.
\end{itemize}
\vspace{-0.5em}
High-stakes examinations demonstrate that such structures are achievable at scale. India's JEE (1.5 million aspirants annually) and China's Gaokao maintain pre-exam security while providing post-hoc transparency: answer keys, scoring rubrics, and formal dispute resolution. This infrastructure is imperfect, but failures trigger investigations and reforms. AI leaderboards currently have none of this.

Why must these institutions be built from inception? Path dependency. Once an organization establishes itself as the de facto evaluator, its rankings become entrenched in procurement criteria, investment decisions, and research benchmarks. Retrofitting governance onto captured infrastructure is far harder than designing it correctly from the start. The Global South has a narrow window: regional leaderboards are emerging now. The choice is whether they emerge with governance or without it.

\textbf{Why regional scale matters for governance.} The argument is not that regional institutions are immune to capture, but that regional scale changes accountability dynamics. First, \textit{proximity}: regional stakeholders can hold regional institutions accountable more directly (a Hindi speaker has more leverage over an Indian institution than a global one). Second, \textit{visibility}: with fewer players, capture is harder to hide; everyone knows who the major actors are. Third, \textit{correction}: reforming a regional institution is more feasible than reforming an entrenched global one. Fourth, \textit{aligned incentives}: regional institutions derive legitimacy from serving regional populations, creating structural pressure toward inclusion that global institutions, responsive primarily to commercial pressure from the Global North, lack.

\section{Case Study: India}
\label{sec:india}

India provides an ideal case study because it exhibits the core dynamics at maximum scale: 1.4 billion people, extreme linguistic diversity (22 scheduled languages, 80+ Hindi dialects), mature regional benchmarks developed by world-class research groups, and significant government investment (\$1.2B IndiaAI Mission). Crucially, India has the technical infrastructure (high-quality benchmarks exist) but lacks the institutional infrastructure for trusted aggregation. This is precisely the gap our position identifies. We use India for depth; Section~\ref{subsec:pattern-validation} validates that the same pattern appears in Africa and the Arabic world.

\subsection{Existing Indic Benchmarks}
\label{subsec:indic-benchmarks}

Indian researchers have developed high-quality benchmarks addressing gaps in global evaluation:

\begin{table}[!htb]
\vspace{-0.5em}
\caption{Existing Indic AI Benchmarks (non-exhaustive)}
\label{tab:benchmarks}
\centering
\small
\begin{tabular}{@{}p{2.2cm}p{1.3cm}p{3.5cm}@{}}
\toprule
\textbf{Benchmark} & \textbf{Type} & \textbf{Coverage} \\
\midrule
IndicSUPERB \citep{javed2023indicsuperb} & ASR & Multi-language evaluation across clean, noisy, telephonic conditions \\
\addlinespace
MILU \citep{ai4bharat2025milu} & LLM & Understanding across 8 domains with Indian context \\
\addlinespace
IndicGenBench \citep{google2024indicgenbench} & LLM & Generation evaluation across multiple Indian languages \\
\addlinespace
LAHAJA \citep{ai4bharat2024lahaja} & Dialect & Hindi regional accent evaluation \\
\addlinespace
Vistaar \citep{ai4bharat2023vistaar} & Domain & News, education, literary, conversational contexts \\
\addlinespace
Svarah \citep{ai4bharat2023svarah} & Accent & Indian English L2 patterns and code-mixing \\
\addlinespace
IISc-MILE \citep{iiscmile2024} & Morphology & Agglutinative evaluation for Tamil and Kannada \\
\bottomrule
\end{tabular}
\end{table}

The technical infrastructure exists. What is missing is \textbf{trusted leaderboard aggregation} with appropriate governance.

\subsection{Stakeholder Consultation Findings}
\label{subsec:consultation}

Nasscom distributed a structured online survey on Indic AI leaderboard design through its stakeholder networks between December 2025 and March 2026, collecting 82 responses from a practitioner-heavy pool.\footnote{Conducted by Nasscom for this position work. Full instrument, demographics, and graphical findings with statistical tests are in Appendices~\ref{app:survey} and \ref{app:results}.} The community judged the regional-leaderboard proposal viable (mean 4.06/5). Two findings stand out. First, every respondent endorsed some form of formal governance; no one chose to leave governance ad hoc. Second, within formal options, 64\% preferred non-government stewardship (32.9\% Nasscom-led, 31.7\% independent non-profit), versus 21\% government-driven and 9.8\% academia-only ($\chi^2 = 26.2$, df=4, $p < 10^{-4}$ vs.\ uniform null). On conflict management, 68\% favored disclosure and recusal over pre-emptive exclusion (95\% Wilson CI [57.6\%, 77.4\%], $p \approx 1.2 \times 10^{-3}$).

\textit{Important limitations:} convenience sample from Nasscom's networks, skewed toward production AI developers (52 of 82, 63\%) and toward respondents already engaged with governance questions. Civil-society and end-user voices are underrepresented; only 4 of 82 respondents identified primarily with government/policy, so public-sector views are likewise thin. We treat this as illustrative triangulation, not representative evidence. The full demographic breakdown is in Appendix~\ref{app:survey}.

\begin{table}[!htb]
\vspace{-0.5em}
\caption{Stakeholder Consultation Key Findings (n=82, Dec 2025 -- Mar 2026)}
\label{tab:survey}
\centering
\small
\begin{tabular}{@{}p{3.2cm}p{1cm}p{2.6cm}@{}}
\toprule
\textbf{Finding} & \textbf{Score} & \textbf{What it tells us} \\
\midrule
Overall viability & 4.06/5 & Community judges the proposal feasible \\
\addlinespace
Endorse formal governance & 100\% & No respondent chose ``no governance'' \\
\addlinespace
Non-government stewardship & 64\% & Nasscom + Independent NP, vs 21\% govt., 9.8\% academia \\
\addlinespace
Conflict management & 68\% & Disclosure/recusal over exclusion \\
\addlinespace
Anti-gaming strategy & 56\% & Dynamic evaluation, even at cost of reproducibility (5 abstentions; 60\% of those who answered) \\
\addlinespace
Evaluation method & 76\% & Hybrid (LLM + human) \\
\addlinespace
Submission cadence & 74\% & Quarterly windows \\
\bottomrule
\end{tabular}
\end{table}

Table~\ref{tab:survey} summarizes key findings. Beyond the headline numbers, the \textit{pattern} of specific preferences is the actionable contribution: a clear tilt toward non-government stewardship paired with disclosure-based conflict management, a community willing to trade some reproducibility for dynamic anti-gaming, and a quarterly refresh cadence. The community's strongest endorsement is for the leaderboard becoming a credible standard (rated 4.33 out of 5); the most cautious dimension is procurement influence (3.63 out of 5), which we treat as evidence that procurement linkage must be deliberately designed rather than assumed.

\subsection{Emerging Regional Infrastructure}
\label{subsec:recent-developments}

Several regional leaderboard initiatives have emerged, validating demand for such infrastructure while illustrating the governance complexity we argue must be addressed early.

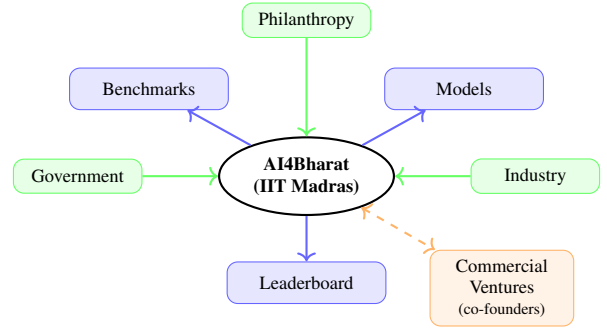
\begin{figure}[!htb]
\centering
\vspace{-0.5em}
\begin{tikzpicture}[
    node distance=0.3cm,
    role/.style={rectangle, draw=blue!60, fill=blue!10, rounded corners, minimum width=2.1cm, minimum height=0.55cm, align=center, font=\scriptsize},
    funding/.style={rectangle, draw=green!60, fill=green!10, rounded corners, minimum width=1.7cm, minimum height=0.45cm, align=center, font=\scriptsize},
    commercial/.style={rectangle, draw=orange!60, fill=orange!10, rounded corners, minimum width=1.9cm, minimum height=0.45cm, align=center, font=\scriptsize},
    center/.style={ellipse, draw=black, thick, minimum width=2cm, minimum height=0.8cm, align=center, font=\scriptsize\bfseries}
]
\node[center] (ai4b) {AI4Bharat\\(IIT Madras)};

\node[role, above left=0.5cm and 0.2cm of ai4b] (bench) {Benchmarks};
\node[role, above right=0.5cm and 0.2cm of ai4b] (models) {Models};
\node[role, below=0.6cm of ai4b] (arena) {Leaderboard};

\node[funding, left=1cm of ai4b] (govt) {Government};
\node[funding, above=1.3cm of ai4b] (phil) {Philanthropy};
\node[funding, right=1cm of ai4b] (google) {Industry};

\node[commercial, below right=0.6cm and 0.8cm of ai4b] (spinout) {Commercial\\Ventures\\{\tiny (co-founders)}};

\draw[->, thick, blue!60] (ai4b) -- (bench);
\draw[->, thick, blue!60] (ai4b) -- (models);
\draw[->, thick, blue!60] (ai4b) -- (arena);
\draw[->, thick, green!60] (govt) -- (ai4b);
\draw[->, thick, green!60] (phil) -- (ai4b);
\draw[->, thick, green!60] (google) -- (ai4b);
\draw[<->, thick, orange!60, dashed] (ai4b) -- (spinout);
\end{tikzpicture}
\vspace{-0.5em}
\caption{AI4Bharat illustrates how regional ecosystems naturally develop: a single organization creates benchmarks, builds models, and operates the leaderboard, with diverse funding sources. This is not criticism but illustration of why governance frameworks become necessary as such entities scale.}
\label{fig:ai4bharat}
\vspace{-1em}
\end{figure}

\textbf{The concentration pattern.} AI4Bharat (IIT Madras) launched the Indic LLM Arena in November 2025 \citep{ai4bharat2025arena}. Figure~\ref{fig:ai4bharat} illustrates the pattern: AI4Bharat creates benchmarks (IndicSUPERB, MILU, LAHAJA), builds models (IndicWhisper, Airavata), and operates the leaderboard, with funding from government, philanthropy, and industry.

Talent concentration is a \textit{feature} of nascent ecosystems, not a flaw unique to any organization. This is precisely why formal COI policies become prerequisites for trust as entities scale. Research shows that \textit{public} disclosure eliminates negative impact on trust \citep{hagan2023editorial}.

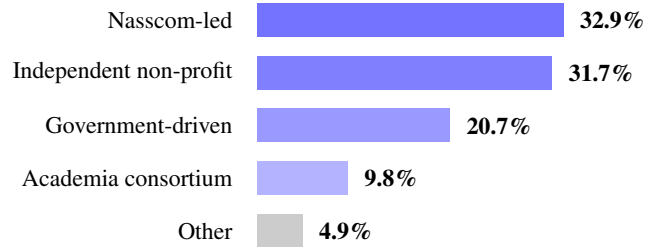
\begin{figure}[!htb]
\centering
\vspace{-0.5em}
\begin{tikzpicture}[scale=1.0]
  \foreach \lbl/\val/\pct/\col [count=\i] in {
    {Nasscom-led}/27/32.9/blue!55,
    {Independent non-profit}/26/31.7/blue!50,
    {Government-driven}/17/20.7/blue!40,
    {Academia consortium}/8/9.8/blue!30,
    {Other}/4/4.9/gray!40} {
    \node[anchor=east, font=\small] at (0,-\i*0.7) {\lbl};
    \fill[\col] (0.2,-\i*0.7-0.22) rectangle ({\val*0.15+0.2},-\i*0.7+0.22);
    \node[anchor=west, font=\small\bfseries] at ({\val*0.15+0.3},-\i*0.7) {\pct\%};
  }
\end{tikzpicture}
\vspace{-0.5em}
\caption{Governance model preference (n=82). All five options are forms of formal governance; no respondent chose ``no governance''. Combined non-government stewardship (Nasscom + Independent non-profit) accounts for 64\% of preferences.}
\label{fig:governance}
\vspace{-1em}
\end{figure}

\subsection{Pattern Validation: Africa and Arabic}
\label{subsec:pattern-validation}

The India case is not unique. \textbf{Africa:} IrokoBench \citep{adelani2025irokobench} provides high-quality evaluation for 16 African languages, yet global leaderboards do not use it. Over 2,000 African languages remain largely neglected \citep{nature2025african}. The governance gap is identical: benchmarks exist, trusted aggregation does not. \textbf{Arabic world:} AlGhafa \citep{alghafa2023} and the Open Arabic LLM Leaderboard \citep{oall2025leaderboard} address Arabic evaluation, but face the same concentrated expertise pattern where benchmark creators also submit models. Arabic, the fifth most spoken language globally, accounts for less than 1\% of training data. In both cases, the technical infrastructure exists; institutional infrastructure does not. India is the detailed case; Africa and Arabic validate the pattern.

\section{Alternative Views}
\label{sec:alternative-views}

We address seven counterarguments to our position.

\subsection{``Regional Leaderboards Risk Capture and Gaming''}
\label{subsec:alt-capture}

\textbf{Argument}: Regional leaderboards may replicate governance failures at smaller scale, with local incumbents capturing governance.

\textbf{Response}: These concerns argue for careful governance design, not against governance. For capture risk: multi-stakeholder boards, term limits, and external audits. For gaming: quarterly refresh with contamination detection.

\subsection{``Fragmentation Will Prevent Interoperability''}
\label{subsec:alt-interoperability}

\textbf{Argument}: Regional leaderboards with different metrics will prevent meaningful cross-regional comparison.

\textbf{Response}: Federation, not unification, is the answer. Cross-regional comparison does not require identical leaderboards; it requires standardized result schemas so that per-language, per-dialect performance can be reported in a common schema and inspected side-by-side, while acknowledging that scores across different benchmarks are not apples-to-apples even when the metric (e.g., WER) is the same. A practical federation has three layers: (1) a shared schema for reporting per-language, per-dialect, per-domain performance (analogous to MLPerf's submission format); (2) a registry of trusted regional leaderboards that conform to baseline governance standards (COI policy, appeals, public methodology); (3) an optional meta-view that surfaces cross-regional comparisons without overriding regional rankings. The MLCommons benchmarking consortium, the W3C web standards process, and the ISO certification ecosystem are existence proofs that federated coordination across independent bodies is feasible. We acknowledge that the protocol specification itself is future work \citep{jrc2024interoperability}, and Appendix~\ref{app:federation} sketches a starting architecture. Transparent diversity serves users better than false uniformity, particularly when uniformity is achieved by excluding the contexts that matter most to the Global South.

\subsection{``Governance Is a Global Problem''}
\label{subsec:alt-global}

\textbf{Argument}: European languages also suffer from governance failures. Isn't this universal?

\textbf{Response}: Yes, but impact is asymmetric. A Basque speaker has ecosystem redundancy: EU regulatory pressure, academic funding, enterprise alternatives. A Santhali or Hausa speaker has no alternative pathway. Market pressure creates accountability for the Global North; governance is the Global South's only mechanism for inclusion.

\subsection{``Organizations Participate in Good Faith, and Regional Advocacy Is Just Nationalism''}
\label{subsec:alt-goodfaith}

\textbf{Argument}: Organizations contribute positively to leaderboards; questioning their involvement is unfair, and regional-leaderboard advocacy is disguised protectionism.

\textbf{Response}: We question governance structures, not intentions. Resource scarcity naturally concentrates expertise across roles (build models, create benchmarks, run leaderboards). This is expected in nascent ecosystems. The question is whether governance exists to manage this concentration. Individual good faith does not substitute for institutional accountability, and network effects create winner-take-all dynamics where building alternative credibility takes years \citep{singh2025leaderboard}. On the nationalism objection: regional leaderboards do not prevent global models from participating. They provide evaluation contexts where regional performance can be fairly assessed. The goal is complementary infrastructure, not replacement.

\subsection{``Funding Risks and Governance Details Unspecified''}
\label{subsec:alt-funding}

\textbf{Argument}: Regional leaderboards require sustained funding, and governance processes remain unspecified.

\textbf{Response}: Evaluation infrastructure is a public good. The question ``who pays?'' is downstream of governance: it cannot be answered without first deciding who bears costs, on whose behalf, and accountable to whom. Saying cost matters more than governance is like saying a budget matters more than a constitution. NIST (publicly funded), MLCommons (multi-stakeholder consortium with industry dues and academic in-kind), and W3C (member dues plus host institutions) are existence proofs. Four sustainability paths are concrete: (a) national AI-mission allocations (India's \$1.2B IndiaAI Mission; similar Indonesian, Brazilian, and African Union initiatives); (b) multi-donor consortia with governance firewalls; (c) industry membership tiers with disclosed contributions and capped voting power; (d) philanthropic anchor funding matched against public co-investment. Procurement linkage strengthens sustainability: ISO certification raises developing-country exports by 44.9\% on average \citep{iso2024exports}. Our respondents are cautious about procurement uptake (3.63 out of 5), which we treat as evidence that procurement linkage requires deliberate design, not a reason to defer governance. Appendix~\ref{app:governance} details the governance framework.

\subsection{``End-User Voice Remains Absent''}
\label{subsec:alt-enduser}

\textbf{Argument}: The paper invokes affected citizens but proposes governance that underweights their representation.

\textbf{Response}: This critique is valid, and argues for stronger governance, not against it. Governance frameworks create the \textit{possibility} of inclusion through user panels and complaint processes. Imperfect representation under transparent governance is improvable; exclusion under no governance is permanent.

\section{Call to Action}
\label{sec:call-to-action}

Our position implies specific actions. A minimum viable regional leaderboard requires: (1) independent governance with a multi-stakeholder board and term limits; (2) a published COI policy; (3) a standardized submission protocol; (4) dispute resolution; (5) interoperability via common result schemas. Sustainable funding models include public funding, industry consortia with governance firewalls, or hybrid approaches. Governance framework details are provided in Appendix~\ref{app:governance}.

\textbf{For nations and funding bodies:} Invest in regional leaderboard infrastructure as a public good with multi-year funding commitments. Aggregate existing benchmarks under transparent governance. Require independence criteria for funded evaluation efforts.

\textbf{For global leaderboard maintainers:} Adopt formal COI policies with disclosure and recusal. Publish testing access policies. Create formal appeals processes. Expand to morphologically appropriate metrics and code-switching evaluation. Include regional benchmarks from the Global South.

\textbf{For researchers and industry:} Recognize regional benchmarks as authoritative for their domains, not secondary to global leaderboards. Require multilingual evaluation for claims of general capability. Support multiple evaluation pathways and contribute to regional benchmark development.

\section{Conclusion}
\label{sec:conclusion}

AI leaderboards are infrastructure. They determine which systems are considered state-of-the-art and enter procurement and investment decisions. When this infrastructure fails to serve a region, the AI ecosystem in that region operates on unreliable signals.

We have argued that AI leaderboards, absent COI policies, independent oversight, and mechanisms for metric evolution, are structurally ill-suited to serving the Global South. The problem is not data. Regional benchmarks exist. The problem is institutional: no governance structure adopts, maintains, and evolves these improvements. Market pressure corrects failures in the Global North; the Global South has no equivalent leverage. Governance is the substitute.

\textbf{Our position is falsifiable.} If global leaderboards demonstrate reliable evaluation with governance that emerging participants trust, regional infrastructure would be unnecessary. The operationalizable conditions are: (1)~\textit{Technical:} report language-family-appropriate metrics, include code-switching evaluation for top-10 multilingual markets, disaggregate by regional accent, and incorporate culturally-grounded knowledge benchmarks; (2)~\textit{Inclusive:} include regional benchmarks (IndicSUPERB, IrokoBench, AlGhafa, SEA-HELM) in standard evaluation suites; (3)~\textit{Adaptive:} maintain formal processes for adopting new metrics as research evolves; (4)~\textit{Governance:} publish explicit COI policies, with more than 50\% of surveyed Global South practitioners rating governance as ``trustworthy''; (5)~\textit{Access:} keep data-access disparity between top-3 commercial providers and median open-weight models at most 2x; (6)~\textit{Impact parity:} governance failures affect Global North and Global South practitioners comparably.

Until these conditions are met, regional leaderboards are not fragmentation of the evaluation landscape but necessary infrastructure. The choice is not between global coordination and regional autonomy but between functional infrastructure and continued underservice of the Global South.

\section*{Acknowledgements}

We thank the ICML 2026 anonymous reviewers and our area chair for substantive engagement during the review and discussion period that materially strengthened this paper, including the suggestions on cross-regional coordination, on funding feasibility, and on the relationship between training-data scarcity and evaluation governance.

\section*{Disclaimer}

The views expressed in this paper are those of the authors in their personal capacity and do not represent the official positions of IIT Kharagpur, Shunya Labs, or Nasscom. This is an argumentative position paper, not an empirical study; the practitioner consultation reported here is illustrative rather than representative, and its limitations are documented in Section~\ref{subsec:consultation} and Appendix~\ref{app:survey}.

\bibliography{references}
\bibliographystyle{icml2026}

\appendix
\section*{\centering Appendix}

\setlength{\intextsep}{5pt plus 1pt minus 1pt}
\setlength{\textfloatsep}{6pt plus 1pt minus 1pt}
\setlength{\floatsep}{10pt plus 2pt minus 2pt}
\setlength{\abovecaptionskip}{3pt}
\setlength{\belowcaptionskip}{2pt}

\section{Survey Instrument and Methodology}
\label{app:survey}

\subsection{Purpose}

The consultation was conducted to gather expert design input for an Indic AI leaderboard. The survey was distributed by Nasscom (the Indian National Association of Software and Service Companies) through its AI stakeholder networks between December 2025 and March 2026. The instrument was developed iteratively with input from industry policy experts and structured around four areas: governance and architecture; technical challenges and solutions; implementation and operations; roadmap and success criteria.

\subsection{Survey Instrument}

All questions permitted multiple selections unless noted as single-select.

\textbf{Q1. Language selection (multi).} Starting set for the Indic AI Leaderboard.

\textbf{Q2. Governance model (single).} Nasscom hosts/governs (hybrid funding, industry-academia advisory); Academia consortium hosts/governs (government funding); Independent non-profit hosts/governs (hybrid funding); Government-driven (MeitY or IndiaAI aegis, PPP funding); Other.

\textbf{Q3. Conflict management (single).} Pre-emptive exclusion vs.\ disclosure/recusal.

\textbf{Q4. ASR datasets (multi).} Vistaar, LAHAJA, IISc-MILE, Svarah, ULCA/Bhashini.

\textbf{Q5. ASR metrics (multi).} WER, CER, KER, plus open write-in.

\textbf{Q6. LLM benchmarks (multi).} MILU, IndicGenBench, IndicMMLU, IndicGLUE, M3LS, IndicXTREME, DRISHTIKON, MultiQ, ParamBench.

\textbf{Q7-Q9. LLM metric families (multi).} Semantic similarity (BERTScore, BLEURT, COMET, BARTScore, PRISM, MoverScore); factuality (FactCC/DAE/QAGS, SummaC); bias/cultural (Stereotype/Bias Score, Cultural Accuracy Score, Indic-Bias, SANSKRITI).

\textbf{Q10. Evaluation method (single).} Human, LLM, or hybrid.

\textbf{Q11. Anti-gaming strategy (single).} Static public, static hidden, dynamic, or adversarial.

\textbf{Q12. Submission requirements (multi).} API documentation, technical specifications, reliability, security.

\textbf{Q13. Submission cadence (single).} Quarterly or half-yearly.

\textbf{Q14. Emerging priorities (multi).} VLMs, multimodal, code generation, agentic AI, domain-specific.

\textbf{Q15. Success criteria (1-5 scale, single).} Used at 2 years; influence on procurement; credibility; ASR/LLM track parity.

\textbf{Q16-Q17. Open-ended.} Additional suggestions.

\textbf{Demographics (multi, with consent).} Stakeholder category, expertise areas, affiliation. No verification of respondent identity.

\subsection{Sampling and Limitations}

Convenience sample drawn from Nasscom's networks. This introduces selection bias toward practitioners with existing engagement in the association's network. Self-reported expertise and affiliations. The final survey window closed at 82 responses. Civil-society and end-user voices remain underrepresented; only 4 of 82 respondents (4.9\%) identified primarily with the ``Government/Policy'' stakeholder category, which limits our ability to disaggregate public-sector views even though the option was offered in the instrument.

\subsection{Note on Survey Evolution (n=58 to n=82)}
\label{app:evolution}

The conference submission cited an interim snapshot of n=58 collected through January 2026. The survey window remained open and continued to collect responses through March 2026, yielding the final n=82 dataset reported in this camera-ready version. All percentages, charts, and headline findings here use the final n=82 data. The change of sample size did not alter the direction of any reported finding. The largest shift between snapshots was the disclosure-versus-exclusion preference (71\% at n=58, 68\% at n=82), which remains a clear majority. We report this evolution transparently in keeping with the governance norms this paper advocates.

\section{Selected Practitioner-Consultation Findings (n=82)}
\label{app:results}
\raggedbottom

\textit{This appendix presents selected findings only. A comprehensive version of the underlying consultation is scheduled for release by Nasscom AI in early July 2026.}

This appendix presents the selected Nasscom consultation findings as bar charts, each paired with a brief statistical test. Unless stated otherwise, percentages are computed over $n=82$, 95\% confidence intervals (CIs) are Wilson intervals, and $p$-values come from exact binomial tests against the stated null or from Pearson $\chi^2$ goodness-of-fit tests against a uniform distribution over the offered options. Likert items (1--5 scale) are tested with one-sample $t$-tests against a neutral mean of 3, using a conservative assumed standard deviation of 1.0.

\textit{Common plot conventions.} All bar charts use raw response counts on the $x$-axis (or $y$-axis for horizontal bars), with $n=82$ unless the item is multi-select (in which case totals exceed 82 and we label the chart accordingly). Where a clear ``null'' alternative exists we annotate the chart with the relevant test statistic.

\subsection{Respondent Demographics}

The respondent pool was practitioner-heavy. Stakeholder category and area-of-expertise were both multi-select questions, so totals exceed 82.

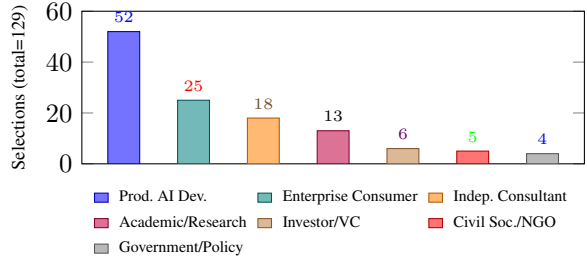
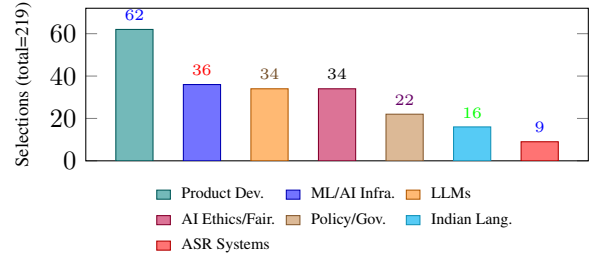
\begin{figure*}[!t]
\centering
\begin{subfigure}{0.48\textwidth}
\centering
\begin{tikzpicture}
\begin{axis}[
    ybar, width=\columnwidth, height=3.6cm,
    bar width=12pt, enlarge x limits=0.10,
    ymin=0, ymax=60, ylabel={\scriptsize Selections (total=129)},
    xtick=\empty,
    nodes near coords, nodes near coords style={font=\tiny},
    label style={font=\scriptsize},
    legend style={
        at={(0.5,-0.10)}, anchor=north,
        font=\tiny, legend columns=3,
        draw=none, fill=none,
        column sep=2pt,
        /tikz/every even column/.append style={column sep=2pt}
    },
    legend cell align={left},
    legend image code/.code={\draw[#1] (0cm,-0.06cm) rectangle (0.18cm,0.06cm);}
]
\addplot+[bar shift=0pt,fill=blue!55,draw=blue!70!black] coordinates {(1,52)};
\addlegendentry{Prod.\ AI Dev.}
\addplot+[bar shift=0pt,fill=teal!55,draw=teal!70!black] coordinates {(2,25)};
\addlegendentry{Enterprise Consumer}
\addplot+[bar shift=0pt,fill=orange!55,draw=orange!70!black] coordinates {(3,18)};
\addlegendentry{Indep.\ Consultant}
\addplot+[bar shift=0pt,fill=purple!55,draw=purple!70!black] coordinates {(4,13)};
\addlegendentry{Academic/Research}
\addplot+[bar shift=0pt,fill=brown!55,draw=brown!70!black] coordinates {(5,6)};
\addlegendentry{Investor/VC}
\addplot+[bar shift=0pt,fill=red!55,draw=red!70!black] coordinates {(6,5)};
\addlegendentry{Civil Soc./NGO}
\addplot+[bar shift=0pt,fill=gray!55,draw=gray!70!black] coordinates {(7,4)};
\addlegendentry{Government/Policy}
\end{axis}
\end{tikzpicture}
\subcaption{Stakeholder categories (total $=$ 129 selections).}
\label{fig:b-demo-stakeholder}
\end{subfigure}\hfill
\begin{subfigure}{0.48\textwidth}
\centering
\begin{tikzpicture}
\begin{axis}[
    ybar, width=\columnwidth, height=3.6cm,
    bar width=14pt, enlarge x limits=0.12,
    ymin=0, ymax=72, ylabel={\scriptsize Selections (total=219)},
    xtick=\empty,
    nodes near coords, nodes near coords style={font=\tiny},
    label style={font=\scriptsize},
    legend style={
        at={(0.5,-0.10)}, anchor=north,
        font=\tiny, legend columns=3,
        draw=none, fill=none,
        column sep=2pt,
        /tikz/every even column/.append style={column sep=2pt}
    },
    legend cell align={left},
    legend image code/.code={\draw[#1] (0cm,-0.06cm) rectangle (0.18cm,0.06cm);}
]
\addplot+[bar shift=0pt,fill=teal!55,draw=teal!70!black] coordinates {(1,62)};
\addlegendentry{Product Dev.}
\addplot+[bar shift=0pt,fill=blue!55,draw=blue!70!black] coordinates {(2,36)};
\addlegendentry{ML/AI Infra.}
\addplot+[bar shift=0pt,fill=orange!55,draw=orange!70!black] coordinates {(3,34)};
\addlegendentry{LLMs}
\addplot+[bar shift=0pt,fill=purple!55,draw=purple!70!black] coordinates {(4,34)};
\addlegendentry{AI Ethics/Fair.}
\addplot+[bar shift=0pt,fill=brown!55,draw=brown!70!black] coordinates {(5,22)};
\addlegendentry{Policy/Gov.}
\addplot+[bar shift=0pt,fill=cyan!55,draw=cyan!70!black] coordinates {(6,16)};
\addlegendentry{Indian Lang.}
\addplot+[bar shift=0pt,fill=red!55,draw=red!70!black] coordinates {(7,9)};
\addlegendentry{ASR Systems}
\end{axis}
\end{tikzpicture}
\subcaption{Expertise areas (total $=$ 219 selections).}
\label{fig:b-demo-expertise}
\end{subfigure}
\caption{Respondent demographics. (a)~Production AI developers (52) are roughly twice the next-largest group, so reported preferences should be read primarily as the views of builders and enterprise deployers; Government/Policy (4) and Civil Society/NGO (5) are the smallest structured groups, and 6 free-text responses (e.g., startup founders, EdTech) bring the total to 129 selections. (b)~ML/AI infrastructure, LLMs, and AI ethics form a tight second tier behind product development, indicating a technically grounded but practitioner-skewed pool; Indian-language linguistics (16) and ASR systems (9) are the smallest structured tiers, and 6 free-text write-ins bring the total to 219.}
\label{fig:b-demographics}
\end{figure*}

\textit{Limitations.} The convenience sample over-represents builders (Production AI Developer + Enterprise AI Consumer + Independent Consultant $=$ 95 of 129 selections, 74\%) and under-represents civil society (5/129, 3.9\%) and government/policy (4/129, 3.1\%). Findings should be read as views of an engaged practitioner community, not a population-representative survey.

\subsection{Language Coverage (Q1)}

\begin{figure*}[!t]
\centering
\begin{subfigure}{0.48\textwidth}
\centering
\begin{tikzpicture}
\begin{axis}[
    xbar, width=\columnwidth, height=5.4cm,
    bar width=5pt, enlarge y limits=0.06,
    xmin=0, xmax=80, xlabel={\scriptsize Respondents endorsing language ($n=82$, multi-select)},
    symbolic y coords={Sanskrit, Urdu, Punjabi, Odia, Assamese, Malayalam, Gujarati,
                       Kannada, Bengali, Marathi, Telugu, Tamil, Hindi, Ind.\ English},
    ytick=data, nodes near coords, nodes near coords align=horizontal,
    nodes near coords style={font=\tiny}, tick label style={font=\tiny},
    label style={font=\scriptsize},
]
\addplot+[fill=orange!55,draw=orange!70!black] coordinates {
    (12,Sanskrit) (10,Urdu) (10,Punjabi) (9,Odia) (8,Assamese)
    (17,Malayalam) (19,Gujarati) (20,Kannada) (22,Bengali) (27,Marathi)
    (34,Telugu) (35,Tamil) (65,Hindi) (71,Ind.\ English)
};
\end{axis}
\end{tikzpicture}
\subcaption{Top tier (counts $\geq 8$).}
\label{fig:b-q1-languages-top}
\end{subfigure}\hfill
\begin{subfigure}{0.48\textwidth}
\centering
\begin{tikzpicture}
\begin{axis}[
    xbar, width=\columnwidth, height=5.4cm,
    bar width=5pt, enlarge y limits=0.06,
    xmin=0, xmax=8, xlabel={\scriptsize Respondents endorsing language ($n=82$, multi-select)},
    symbolic y coords={Bodo, Dogri, Nepali, Maithili, Manipuri, Santali, Sindhi,
                       Kashmiri, Konkani},
    ytick=data, nodes near coords, nodes near coords align=horizontal,
    nodes near coords style={font=\tiny}, tick label style={font=\tiny},
    label style={font=\scriptsize},
]
\addplot+[fill=orange!35,draw=orange!60!black] coordinates {
    (2,Bodo) (2,Dogri) (2,Nepali) (3,Maithili) (3,Manipuri) (3,Santali)
    (3,Sindhi) (4,Kashmiri) (5,Konkani)
};
\end{axis}
\end{tikzpicture}
\subcaption{Long tail (counts $\leq 5$).}
\label{fig:b-q1-languages-tail}
\end{subfigure}
\caption{Language coverage priorities. Indian English (71) and Hindi (65) are clear anchors. Six languages clear the 22-respondent ($>$25\%) threshold: Indian English, Hindi, Tamil, Telugu, Marathi, Bengali. The long tail (17 of 23 listed languages with $<$25\% endorsement) signals demand for phased but broad eventual coverage. Sanskrit (12) over-indexes relative to its spoken-population share, reflecting the academic sub-pool.}
\label{fig:b-q1-languages}
\end{figure*}

\subsection{Governance Preference (Q2)}

\begin{figure}[H]
\centering
\begin{tikzpicture}
\begin{axis}[
    ybar, width=0.95\columnwidth, height=4.5cm,
    bar width=12pt, enlarge x limits=0.12,
    ymin=0, ymax=36, ylabel={\scriptsize Respondents ($n=82$)},
    symbolic x coords={Nasscom-led, Indep.\ NP, Government, Academia, Other},
    xtick=data, nodes near coords, nodes near coords style={font=\tiny},
    tick label style={font=\tiny}, x tick label style={font=\tiny,rotate=20,anchor=east},
    label style={font=\scriptsize},
]
\addplot+[fill=purple!45,draw=purple!60!black] coordinates {
    (Nasscom-led,27) (Indep.\ NP,26) (Government,17) (Academia,8) (Other,4)
};
\draw[dashed,red!70!black,thick] (axis cs:Nasscom-led,16.4) -- (axis cs:Other,16.4);
\node[font=\tiny,red!70!black,anchor=north east] at (rel axis cs:0.98,0.95) {uniform null $=$ 16.4};
\end{axis}
\end{tikzpicture}
\caption{Governance model preference. Non-government stewardship (Nasscom-led $+$ independent non-profit) $=$ 53/82, 64.6\% (95\% CI [53.8\%, 74.1\%]). Distribution is significantly non-uniform: $\chi^2 = 26.2$, df $= 4$, $p = 2.9 \times 10^{-5}$. All four structured options are formal-governance models; no respondent chose ``no governance''. The two community-anchored options jointly carry the modal weight.}
\label{fig:b-q2-governance}
\end{figure}
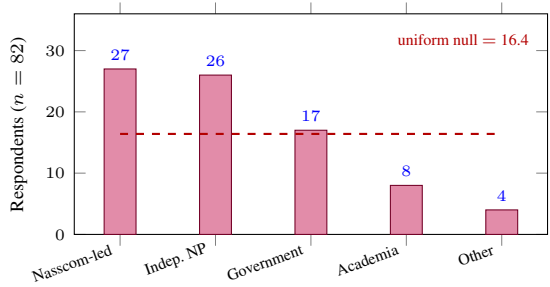

\subsection{Conflict Management (Q3)}

\begin{figure}[H]
\centering
\begin{tikzpicture}
\begin{axis}[
    ybar, width=0.85\columnwidth, height=4cm,
    bar width=22pt, enlarge x limits=0.4,
    ymin=0, ymax=70, ylabel={\scriptsize Respondents ($n=82$)},
    symbolic x coords={Disclosure/Recusal, Pre-emptive Exclusion},
    xtick=data, nodes near coords, nodes near coords style={font=\tiny},
    tick label style={font=\tiny}, label style={font=\scriptsize},
]
\addplot+[fill=cyan!50,draw=cyan!60!black] coordinates {
    (Disclosure/Recusal,56) (Pre-emptive Exclusion,26)
};
\end{axis}
\end{tikzpicture}
\caption{Conflict-of-interest management preference. Disclosure/recusal $=$ 56/82, 68.3\% (95\% CI [57.6\%, 77.4\%]). Two-sided binomial test vs.\ a 50/50 null: $p = 1.2 \times 10^{-3}$. The community prefers managed participation over exclusion.}
\label{fig:b-q3-coi}
\end{figure}
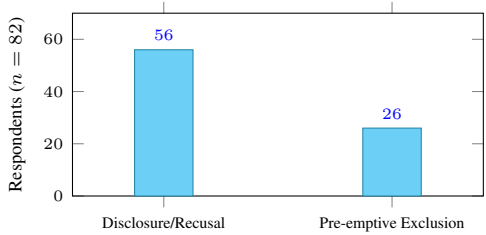

\subsection{ASR Datasets and Metrics (Q4, Q5)}

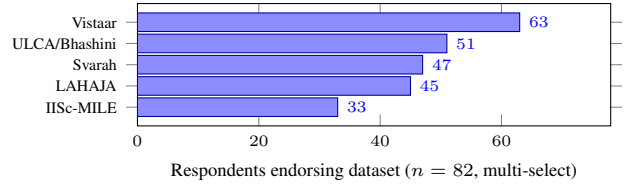
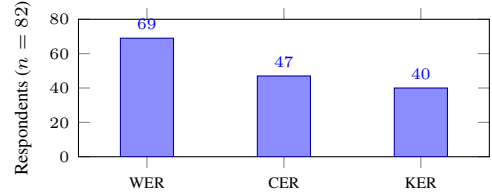
\begin{figure}[H]
\centering
\begin{subfigure}{\columnwidth}
\centering
\begin{tikzpicture}
\begin{axis}[
    xbar, width=0.95\columnwidth, height=3.2cm,
    bar width=7pt, enlarge y limits=0.22,
    xmin=0, xmax=78, xlabel={\scriptsize Respondents endorsing dataset ($n=82$, multi-select)},
    symbolic y coords={IISc-MILE, LAHAJA, Svarah, ULCA/Bhashini, Vistaar},
    ytick=data, nodes near coords, nodes near coords align=horizontal,
    nodes near coords style={font=\tiny}, tick label style={font=\tiny},
    label style={font=\scriptsize},
]
\addplot+[fill=blue!45,draw=blue!60!black] coordinates {
    (33,IISc-MILE) (45,LAHAJA) (47,Svarah) (51,ULCA/Bhashini) (63,Vistaar)
};
\end{axis}
\end{tikzpicture}
\subcaption{Datasets: Vistaar (63/82, 77\%) anchors; four clear 50\%.}
\label{fig:b-q4-asr-data}
\end{subfigure}

\medskip
\begin{subfigure}{\columnwidth}
\centering
\begin{tikzpicture}
\begin{axis}[
    ybar, width=0.85\columnwidth, height=3.4cm,
    bar width=20pt, enlarge x limits=0.25,
    ymin=0, ymax=80, ylabel={\scriptsize Respondents ($n=82$)},
    symbolic x coords={WER, CER, KER},
    xtick=data, nodes near coords, nodes near coords style={font=\tiny},
    tick label style={font=\tiny}, label style={font=\scriptsize},
]
\addplot+[fill=blue!45,draw=blue!60!black] coordinates {
    (WER,69) (CER,47) (KER,40)
};
\end{axis}
\end{tikzpicture}
\subcaption{Metrics: WER (84\%), CER (57\%), KER (49\%).}
\label{fig:b-q5-asr-metric}
\end{subfigure}
\caption{ASR datasets and metrics. (a)~Vistaar (63/82, 77\%) is the natural anchor; four datasets clear 50\% endorsement, indicating expectation of a multi-dataset ASR suite rather than a single benchmark. (b)~WER is dominant; CER and KER are widely endorsed as complements rather than alternatives. Ten open-text write-ins requested semantic-similarity and token-error variants, signalling that standard rate-based metrics are necessary but not sufficient for Indic-language nuance.}
\label{fig:b-asr-datasets-metrics}
\end{figure}

\subsection{LLM Datasets (Q6) and Metric Families (Q7--Q9)}

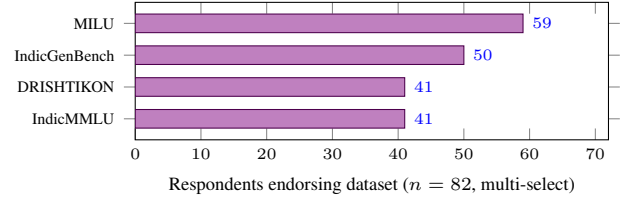
\begin{figure}[H]
\centering
\begin{tikzpicture}
\begin{axis}[
    xbar, width=0.95\columnwidth, height=3.4cm,
    bar width=7pt, enlarge y limits=0.22,
    xmin=0, xmax=72, xlabel={\scriptsize Respondents endorsing dataset ($n=82$, multi-select)},
    symbolic y coords={IndicMMLU, DRISHTIKON, IndicGenBench, MILU},
    ytick=data, nodes near coords, nodes near coords align=horizontal,
    nodes near coords style={font=\tiny}, tick label style={font=\tiny},
    label style={font=\scriptsize},
]
\addplot+[fill=violet!50,draw=violet!60!black] coordinates {
    (41,IndicMMLU) (41,DRISHTIKON) (50,IndicGenBench) (59,MILU)
};
\end{axis}
\end{tikzpicture}
\caption{LLM datasets. MILU (59/82, 72\%) leads; no dataset clears 75\% endorsement, confirming that no single dataset covers language breadth, cultural depth, domain knowledge, and generation capability simultaneously. A composite suite is expected.}
\label{fig:b-q6-llm-data}
\end{figure}

\begin{figure}[H]
\centering
\begin{tikzpicture}
\begin{axis}[
    xbar, width=0.95\columnwidth, height=4.4cm,
    bar width=6pt, enlarge y limits=0.10,
    xmin=0, xmax=78, xlabel={\scriptsize Respondents ($n=82$, multi-select)},
    symbolic y coords={MoverScore, PRISM, BARTScore, BLEURT, COMET, BERTScore},
    ytick=data, nodes near coords, nodes near coords align=horizontal,
    nodes near coords style={font=\tiny}, tick label style={font=\tiny},
    label style={font=\scriptsize},
]
\addplot+[fill=violet!50,draw=violet!60!black] coordinates {
    (27,MoverScore) (30,PRISM) (33,BARTScore) (34,BLEURT) (35,COMET) (63,BERTScore)
};
\end{axis}
\end{tikzpicture}
\caption{Semantic-similarity metrics for LLMs. BERTScore (77\%) dominates; five secondary metrics cluster tightly between 33\% and 43\%, indicating expectation of a multi-metric semantic suite rather than a single canonical score.}
\label{fig:b-q7-llm-semantic}
\end{figure}
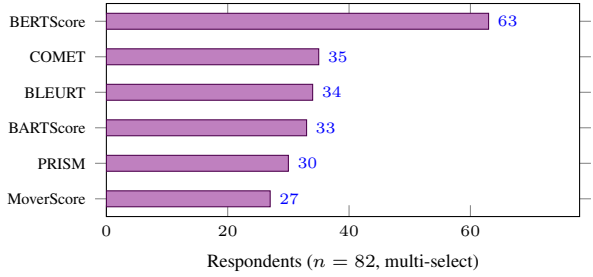

\begin{figure}[H]
\centering
\begin{tikzpicture}
\begin{axis}[
    ybar, width=0.85\columnwidth, height=3.8cm,
    bar width=20pt, enlarge x limits=0.3,
    ymin=0, ymax=70, ylabel={\scriptsize Respondents ($n=82$)},
    symbolic x coords={Stereotype/Bias, Cultural Accuracy},
    xtick=data, nodes near coords, nodes near coords style={font=\tiny},
    tick label style={font=\tiny}, label style={font=\scriptsize},
]
\addplot+[fill=red!45,draw=red!60!black] coordinates {
    (Stereotype/Bias,59) (Cultural Accuracy,56)
};
\end{axis}
\end{tikzpicture}
\caption{Bias and cultural-fairness metrics. The two scores are statistically indistinguishable (Stereotype/Bias 72\%, Cultural Accuracy 68\%; difference of proportions $p = 0.62$). Both are treated as essential, reflecting the community's view that Indic AI evaluation must explicitly test for social fairness, not only task accuracy. Factuality items (FactCC/DAE/QAGS family, 67/82, 82\%) are the clear leader within that separate question.}
\label{fig:b-q9-bias}
\end{figure}
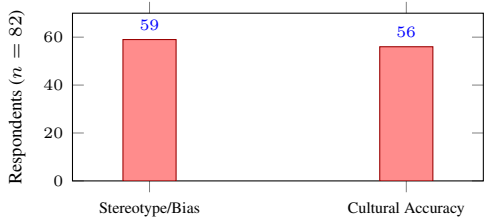

\subsection{Evaluation Method (Q10)}

\begin{figure}[H]
\centering
\begin{tikzpicture}
\begin{axis}[
    ybar, width=0.95\columnwidth, height=4.2cm,
    bar width=18pt, enlarge x limits=0.18,
    ymin=0, ymax=72, ylabel={\scriptsize Respondents ($n=82$)},
    symbolic x coords={Hybrid LLM+Human, LLM Judges, Human Judges},
    xtick=data, nodes near coords, nodes near coords style={font=\tiny},
    tick label style={font=\tiny}, x tick label style={font=\tiny},
    label style={font=\scriptsize},
]
\addplot+[fill=green!45!black,draw=green!50!black] coordinates {
    (Hybrid LLM+Human,62) (LLM Judges,12) (Human Judges,8)
};
\draw[dashed,red!70!black,thick] (axis cs:Hybrid LLM+Human,27.33) -- (axis cs:Human Judges,27.33);
\node[font=\tiny,red!70!black,anchor=north east] at (rel axis cs:0.98,0.95) {uniform null $=$ 27.3};
\end{axis}
\end{tikzpicture}
\caption{Evaluation method preference. Hybrid (LLM+human) $=$ 62/82, 75.6\% (95\% CI [65.3\%, 83.6\%]). Test against a uniform 3-option null: $\chi^2 = 66.2$, df $= 2$, $p < 10^{-14}$. Hybrid evaluation reflects awareness of both LLM-judge biases and human-judge scalability limits.}
\label{fig:b-q10-eval}
\end{figure}
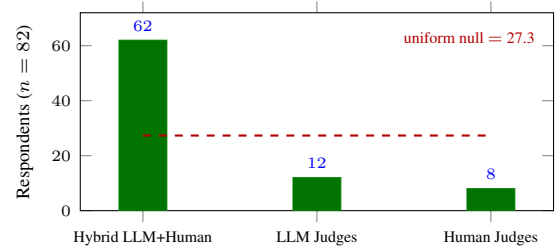

\subsection{Anti-Gaming Strategy (Q11)}

\begin{figure}[H]
\centering
\begin{tikzpicture}
\begin{axis}[
    ybar, width=0.95\columnwidth, height=4.2cm,
    bar width=12pt, enlarge x limits=0.10,
    ymin=0, ymax=55, ylabel={\scriptsize Respondents ($n=77$ answered)},
    symbolic x coords={Dynamic, Static Public, Adversarial, Static Hidden},
    xtick=data, nodes near coords, nodes near coords style={font=\tiny},
    tick label style={font=\tiny}, x tick label style={font=\tiny,rotate=15,anchor=east},
    label style={font=\scriptsize},
]
\addplot+[fill=brown!55,draw=brown!60!black] coordinates {
    (Dynamic,46) (Static Public,16) (Adversarial,11) (Static Hidden,4)
};
\draw[dashed,red!70!black,thick] (axis cs:Dynamic,19.25) -- (axis cs:Static Hidden,19.25);
\node[font=\tiny,red!70!black,anchor=north east] at (rel axis cs:0.98,0.95) {uniform null $=$ 19.25};
\end{axis}
\end{tikzpicture}
\caption{Anti-gaming strategy (5 of 82 respondents abstained). Dynamic evaluation $=$ 46/82, 56.1\% (95\% CI [45.3\%, 66.3\%]); among the 77 who answered, Dynamic was 59.7\%. Test against a uniform 4-option null over the 77 answered responses: $\chi^2 = 53.3$, df $= 3$, $p < 10^{-10}$. Even at the cost of strict reproducibility, the community prefers leaderboards that can rotate test items to resist memorization and overfitting.}
\label{fig:b-q11-antigaming}
\end{figure}
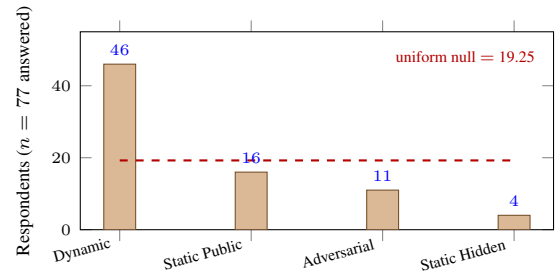

\subsection{Submission Requirements and Cadence (Q12, Q13)}

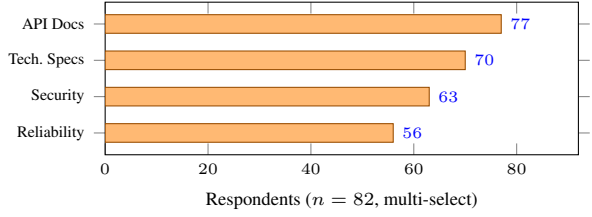
\begin{figure}[H]
\centering
\begin{tikzpicture}
\begin{axis}[
    xbar, width=0.95\columnwidth, height=3.6cm,
    bar width=7pt, enlarge y limits=0.20,
    xmin=0, xmax=92, xlabel={\scriptsize Respondents ($n=82$, multi-select)},
    symbolic y coords={Reliability, Security, Tech.\ Specs, API Docs},
    ytick=data, nodes near coords, nodes near coords align=horizontal,
    nodes near coords style={font=\tiny}, tick label style={font=\tiny},
    label style={font=\scriptsize},
]
\addplot+[fill=orange!55,draw=orange!60!black] coordinates {
    (56,Reliability) (63,Security) (70,Tech.\ Specs) (77,API Docs)
};
\end{axis}
\end{tikzpicture}
\caption{Submission requirements. All four standards-style requirements clear 68\% endorsement (lowest is Reliability at 56/82, 68\%; highest is API Documentation at 94\%), indicating broad alignment that submissions should meet production-grade engineering standards before evaluation.}
\label{fig:b-q12-reqs}
\end{figure}

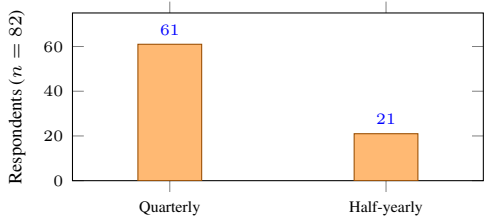
\begin{figure}[H]
\centering
\begin{tikzpicture}
\begin{axis}[
    ybar, width=0.85\columnwidth, height=3.8cm,
    bar width=24pt, enlarge x limits=0.45,
    ymin=0, ymax=75, ylabel={\scriptsize Respondents ($n=82$)},
    symbolic x coords={Quarterly, Half-yearly},
    xtick=data, nodes near coords, nodes near coords style={font=\tiny},
    tick label style={font=\tiny}, label style={font=\scriptsize},
]
\addplot+[fill=orange!55,draw=orange!60!black] coordinates {
    (Quarterly,61) (Half-yearly,21)
};
\end{axis}
\end{tikzpicture}
\caption{Submission cadence. Quarterly windows $=$ 61/82, 74.4\% (95\% CI [64.0\%, 82.6\%]). Two-sided binomial test vs.\ a 50/50 null: $p = 1.1 \times 10^{-5}$. Three of four respondents want four submission windows per year, balancing freshness against operational burden.}
\label{fig:b-q13-cadence}
\end{figure}

\subsection{Emerging Technology Priorities (Q14)}

\begin{figure}[H]
\centering
\resizebox{\columnwidth}{!}{\begin{tikzpicture}
\begin{axis}[
    ybar, width=0.92\columnwidth, height=3.6cm,
    bar width=14pt, enlarge x limits=0.12,
    ymin=0, ymax=78, ylabel={\scriptsize Respondents ($n=82$)},
    xtick=\empty,
    nodes near coords, nodes near coords style={font=\tiny},
    label style={font=\scriptsize},
    legend style={
        at={(0.5,-0.08)}, anchor=north,
        font=\tiny, legend columns=3,
        draw=none, fill=none,
        column sep=2pt,
        /tikz/every even column/.append style={column sep=2pt}
    },
    legend cell align={left},
    legend image code/.code={\draw[#1] (0cm,-0.06cm) rectangle (0.18cm,0.06cm);}
]
\addplot+[bar shift=0pt,fill=magenta!55,draw=magenta!70!black] coordinates {(1,66)};
\addlegendentry{Specialized Dom.}
\addplot+[bar shift=0pt,fill=blue!55,draw=blue!70!black] coordinates {(2,61)};
\addlegendentry{Multimodal}
\addplot+[bar shift=0pt,fill=teal!55,draw=teal!70!black] coordinates {(3,60)};
\addlegendentry{Agentic AI}
\addplot+[bar shift=0pt,fill=orange!55,draw=orange!70!black] coordinates {(4,59)};
\addlegendentry{Visual Language}
\addplot+[bar shift=0pt,fill=purple!55,draw=purple!70!black] coordinates {(5,33)};
\addlegendentry{Code Gen.}
\end{axis}
\end{tikzpicture}}
\caption{Emerging-technology priorities. Specialized domains (healthcare, legal, agriculture, education) $=$ 66/82, 80.5\% (95\% CI [70.6\%, 87.6\%]; vs.\ 50\% null: $p = 2.3 \times 10^{-8}$). Multimodal (74\%) and agentic (73\%) are statistically indistinguishable from each other. Code generation (40\%) trails noticeably, suggesting the community sees Indic-language code as a secondary near-term priority.}
\label{fig:b-q14-tech}
\end{figure}
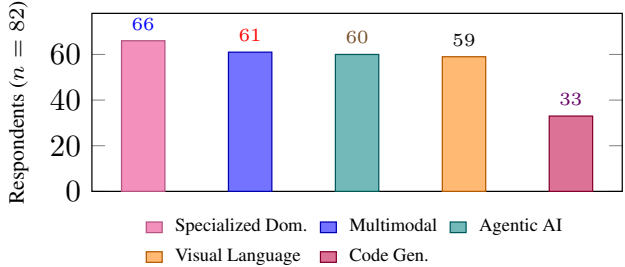

\subsection{Success Criteria (Q15, 1--5 Scale)}

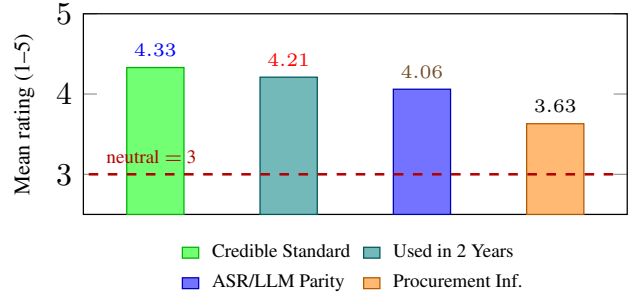
\begin{figure}[H]
\centering
\resizebox{\columnwidth}{!}{\begin{tikzpicture}
\begin{axis}[
    ybar, width=0.92\columnwidth, height=3.8cm,
    bar width=18pt, enlarge x limits=0.18,
    ymin=2.5, ymax=5.0, ylabel={\scriptsize Mean rating (1--5)},
    xtick=\empty,
    nodes near coords={\pgfmathprintnumber[fixed,precision=2]{\pgfplotspointmeta}},
    nodes near coords style={font=\tiny}, label style={font=\scriptsize},
    legend style={
        at={(0.5,-0.08)}, anchor=north,
        font=\tiny, legend columns=2,
        draw=none, fill=none,
        column sep=2pt,
        /tikz/every even column/.append style={column sep=2pt}
    },
    legend cell align={left},
    legend image code/.code={\draw[#1] (0cm,-0.06cm) rectangle (0.18cm,0.06cm);}
]
\addplot+[bar shift=0pt,fill=green!55,draw=green!70!black] coordinates {(1,4.33)};
\addlegendentry{Credible Standard}
\addplot+[bar shift=0pt,fill=teal!55,draw=teal!70!black] coordinates {(2,4.21)};
\addlegendentry{Used in 2 Years}
\addplot+[bar shift=0pt,fill=blue!55,draw=blue!70!black] coordinates {(3,4.06)};
\addlegendentry{ASR/LLM Parity}
\addplot+[bar shift=0pt,fill=orange!55,draw=orange!70!black] coordinates {(4,3.63)};
\addlegendentry{Procurement Inf.}
\draw[dashed,red!70!black,thick] (axis cs:0.5,3) -- (axis cs:4.5,3);
\node[font=\tiny,red!70!black,anchor=south west] at (axis cs:0.55,3) {neutral $=$ 3};
\end{axis}
\end{tikzpicture}}
\caption{Success criteria, mean Likert rating ($n=82$, neutral $=$ 3). All four items are significantly above neutral (one-sample $t$-tests with assumed SD$=$1): Credible Standard $t=12.1$, $p<10^{-15}$; Used in 2 Years $t=11.0$, $p<10^{-15}$; ASR/LLM Parity $t=9.6$, $p<10^{-15}$; Procurement Influence $t=5.7$, $p<10^{-7}$. The community is most confident in trust building (4.33) and most cautious on government procurement uptake (3.63), suggesting that policy linkage is a follower variable rather than a precondition.}
\label{fig:b-q15-success}
\end{figure}

\subsection{What the Statistical Inference Adds}

The bar charts above show that the headline percentages reported in Section~\ref{subsec:consultation} are not artifacts of small samples or balanced splits. For every multi-option question, the response distribution is significantly non-uniform at $p < 10^{-4}$ or better, with the modal choice having a 95\% CI that does not include the chance level. For every dichotomous question, the majority preference is significant at $p < 10^{-3}$. For every Likert success criterion, the mean is significantly above neutral at $p < 10^{-7}$. The consultation cannot establish population-level claims (the sample is a convenience pool, not a probability sample), but within that pool the preferences are sharp, consistent, and unlikely under any plausible chance-only model.

\flushbottom

\section{Illustrative Governance Framework}
\label{app:governance}

This appendix presents one possible instantiation of governance structures, not a prescriptive specification. Regional implementations should adapt these elements to local institutional contexts, legal frameworks, and stakeholder needs.

\begin{table}[!htb]
\vspace{-0.5em}
\caption{Board Composition and Terms}
\centering
\small
\begin{tabular}{@{}p{2.2cm}p{0.9cm}p{3.0cm}@{}}
\toprule
\textbf{Stakeholder} & \textbf{Share} & \textbf{Constraints} \\
\midrule
Academia & 30\% & Max 3 seats \\
Industry & 30\% & No company $>$1 seat \\
Government & 20\% & Observer or voting \\
Civil Society & 10\% & 1 seat \\
International & 10\% & External perspective \\
\midrule
\multicolumn{3}{@{}p{6.1cm}@{}}{\footnotesize\textit{Terms: 3 years (max 2 consecutive); Chair rotates; 2-year cooling-off period.}} \\
\bottomrule
\end{tabular}
\vspace{-1em}
\end{table}

\begin{table}[!htb]
\vspace{-0.5em}
\caption{Conflict of Interest Policy}
\centering
\small
\begin{tabular}{@{}p{1.8cm}p{4.4cm}@{}}
\toprule
\textbf{Category} & \textbf{Requirements} \\
\midrule
Disclosure & Annual financial interests; submitted models; funding relationships; public register \\
Recusal & Decisions affecting own interests; methodology changes; affiliated disputes \\
Prohibited & Board cannot submit models; staff cannot hold equity; no consulting with evaluated entities \\
\bottomrule
\end{tabular}
\end{table}

\section{Cross-Modal Governance Gap}
\label{app:crossmodal}

The body (Section~\ref{subsec:technical}) summarizes evidence that the governance gap extends beyond language models. This appendix documents the cases in more detail.

\textbf{Medical diagnostics.} A 2025 study of vision-language models in chest X-ray diagnosis found that models consistently underdiagnose marginalized groups, with the highest error rates for Black female patients \citep{yang2025chestxray}. In cardiology, a study of 4{,}975 first-time heart-attack patients in India found that the widely used Western cardiovascular risk models (Framingham, ACC/AHA ASCVD, WHO charts) misclassified nearly 80\% of patients as low or moderate risk, despite the patients having presented with acute infarction \citep{gupta2026cardio}. The models were never re-validated against South Asian physiology and lipid profiles before being adopted as clinical reference points.

\textbf{Agriculture.} The PlantVillage crop-disease detection model reported 99.35\% accuracy on its North American lab benchmark \citep{mohanty2016plantvillage}. When the same model class was deployed for cassava disease detection in Tanzania, accuracy fell to 49\% \citep{mrisho2020cassava}. The drop reflects field-versus-lab image conditions, regional disease variants, and crop varieties absent from training data. No governance mechanism required validation on the deployment population before the model was promoted to smallholder farmers.

\textbf{Weather prediction.} Modern AI weather models such as Pangu-Weather, GraphCast, and FourCastNet are trained on the ECMWF ERA5 reanalysis, which itself is constructed from a global observation network in which Africa has roughly one-eighth of the WMO-recommended surface-station density \citep{wmo2024stations}. AI weather models inherit this observational bias, producing higher forecast errors over tropical regions \citep{mozaffari2026weather}. Communities most exposed to climate volatility are the least well served by the systems being deployed to manage it.

\textbf{The common pattern.} In every case the technical fix is known (collect more local data, run domain-adapted fine-tuning, re-validate against local populations). The bottleneck is institutional: no governance mechanism mandates validation on the affected populations before deployment, and no appeals process exists when deployed systems fail. A better maker is not a substitute for a checker.

\section{Federation Architecture Sketch}
\label{app:federation}

Section~\ref{subsec:alt-interoperability} argues that federation, not unification, is the answer to cross-regional comparability. This appendix sketches a starting architecture.

\textbf{Layer 1: Shared result schema.} A common JSON schema for reporting per-language, per-dialect, per-domain, and per-task performance, analogous to the MLPerf submission format. The schema is descriptive (``Model X scored 94\% CER on Hindi LAHAJA dev split, run on date Y, hardware Z, prompt template P''), not prescriptive about which metrics to optimize. Regional leaderboards remain free to weight metrics differently in their headline rankings.

\textbf{Layer 2: Trusted-leaderboard registry.} A lightweight registry of regional leaderboards that conform to baseline governance standards: a published COI policy, a documented appeals process, a public methodology, and an independent fairness-audit panel. Conformance is self-declared with periodic peer review (analogous to ICANN's accreditation of registrars or ISO's accreditation of national standards bodies).

\textbf{Layer 3: Optional meta-view.} A meta-leaderboard, hosted by a neutral body (such as MLCommons or a UN-affiliated entity), surfaces cross-regional comparisons in the shared schema without overriding regional rankings. The meta-view exposes capability claims to multilingual scrutiny without imposing a global ranking.

\textbf{Why this works.} The three layers separate three concerns that the current ecosystem conflates: (1) what to measure (regional autonomy), (2) how to report measurements (shared schema), and (3) who can claim trust (registry). Federation across MLCommons, NIST, W3C, and ISO operates on similar principles. Protocol specification is future work and beyond the scope of this position.

\end{document}